\documentclass[sigconf]{acmart}

\usepackage{booktabs} % For formal tables
\usepackage[ruled]{algorithm2e} % For algorithms 
\usepackage{algpseudocode} 
\usepackage{graphicx}
\usepackage{subcaption}
\usepackage{tabularx}
\usepackage{epsfig}  
\usepackage{xcolor}
\usepackage{amsthm}
\usepackage{multirow}
\usepackage{amssymb}
\usepackage{appendix}
\usepackage{xcolor}
\usepackage{pdfpages}
\usepackage{url}
\usepackage{fancyhdr}
\graphicspath{{./figs/}{./plots}{./images/}}
\newcommand{\eat}[1]{}

\DeclareMathOperator*{\diag}{diag\,}
\DeclareMathOperator*{\simfunc}{sim\,}

\acmYear{2019}
\copyrightyear{2019}
\acmPrice{15.00}

\begin{document}
\title[]{Dynamic Demand Prediction for Expanding Electric Vehicle Sharing Systems}
\author{Man Luo$^1$, Hongkai Wen$^{1}$, Yi Luo$^2$, Bowen Du$^1$, Konstantin Klemmer$^{1}$ and Hongming Zhu$^2$}
% \thanks{Correspondence author.}
\affiliation{$^1$Department of Computer Science, University of Warwick, UK}
\affiliation{$^2$School of Software Engineering, Tongji University, China}
% \affiliation{$^3$The Alan Turing Institute, UK}
\affiliation{\{m.luo.1, hongkai.wen, b.du, k.klemmer\}@warwick.ac.uk, \{1731530, zhu\_hongming\}@tongji.edu.cn}

\renewcommand{\shortauthors}{M. Luo et al.}

%!TEX root = main.tex

\begin{abstract}
Electric Vehicle (EV) sharing systems have recently experienced unprecedented growth across the globe. Many car sharing service providers as well as automobile manufacturers are entering this competition by expanding both their EV fleets and renting/returning station networks, aiming to seize a share of the market and bring car sharing to the zero emissions level. During their fast expansion, one fundamental determinant for success is the capability of dynamically predicting the demand of stations. In this paper we propose a novel demand prediction approach, which is able to model the dynamics of the system and predict demand accordingly. We use a local temporal encoding process to handle the available historical data at individual stations, and a spatial encoding process to take correlations between stations into account with graph convolutional neural networks. The encoded features are fed to a prediction network, which forecasts both the long-term expected demand of the stations. We evaluate the proposed approach on real-world data collected from a major EV sharing platform. Experimental results demonstrate that our approach significantly outperforms the state of the art.
\end{abstract}

% \keywords{Electric Vehicle Sharing; Dynamic Demand Prediction; Expansion}

\maketitle

% \renewcommand{\thefootnote}{\fnsymbol{footnote}}
%!TEX root = main.tex
\section{Introduction}

Car sharing services have long been recognised as an environmentally friendly mobility option, reducing vehicles on the road while cutting out unnecessary CO$_2$ emissions. With the recent advances in battery technologies, a new generation of car sharing services is going one step further, by offering full electric vehicle (EV) fleets with fast expanding infrastructures in major cities, e.g. Bluecity~\footnote{\url{https://www.blue-city.co.uk}} in London, WeShare~\footnote{\url{https://www.volkswagenag.com/en/news/2018/08/VW_Brand_We_Share.html}} in Berlin, and BlueSG~\footnote{\url{https://www.bluesg.com.sg}} in Singapore. Traditional car sharing providers have also started to populate their EV fleets, e.g., ZipCar seeks to provide over 9,000 full electric vehicles across London by 2025~\footnote{\url{https://www.zipcar.co.uk/electric}}. According to a recent study~\cite{Shaheen:2018}, the global market of EV sharing services is poised for much faster growth in the near future, due to the incentives and regulations put in place by governments across the world to encourage overall EV usages. 

%% Need dynamic prediction : both expected and instant.
Despite their increased popularity, the practicality and utility of EV sharing systems still rely heavily on the infrastructure at renting/returning stations. In particular, for systems with the need to rapidly expand their station networks, it is paramount to be able to dynamically predict the accurate demand as or even before implementing any expansion strategy. This is not only the key for the stakeholders to make informed decisions as to where and when to deploy new stations or close the poorly performing ones, but also of great importance to the effective operation of currently used stations, since understanding the potential impact of proposed expansion to their demand can provide valuable insights on a number of vital tasks such as scheduling and rebalancing.

\begin{figure*}[t]
\centering
\begin{subfigure}{0.5\columnwidth}
\includegraphics[width=\textwidth]{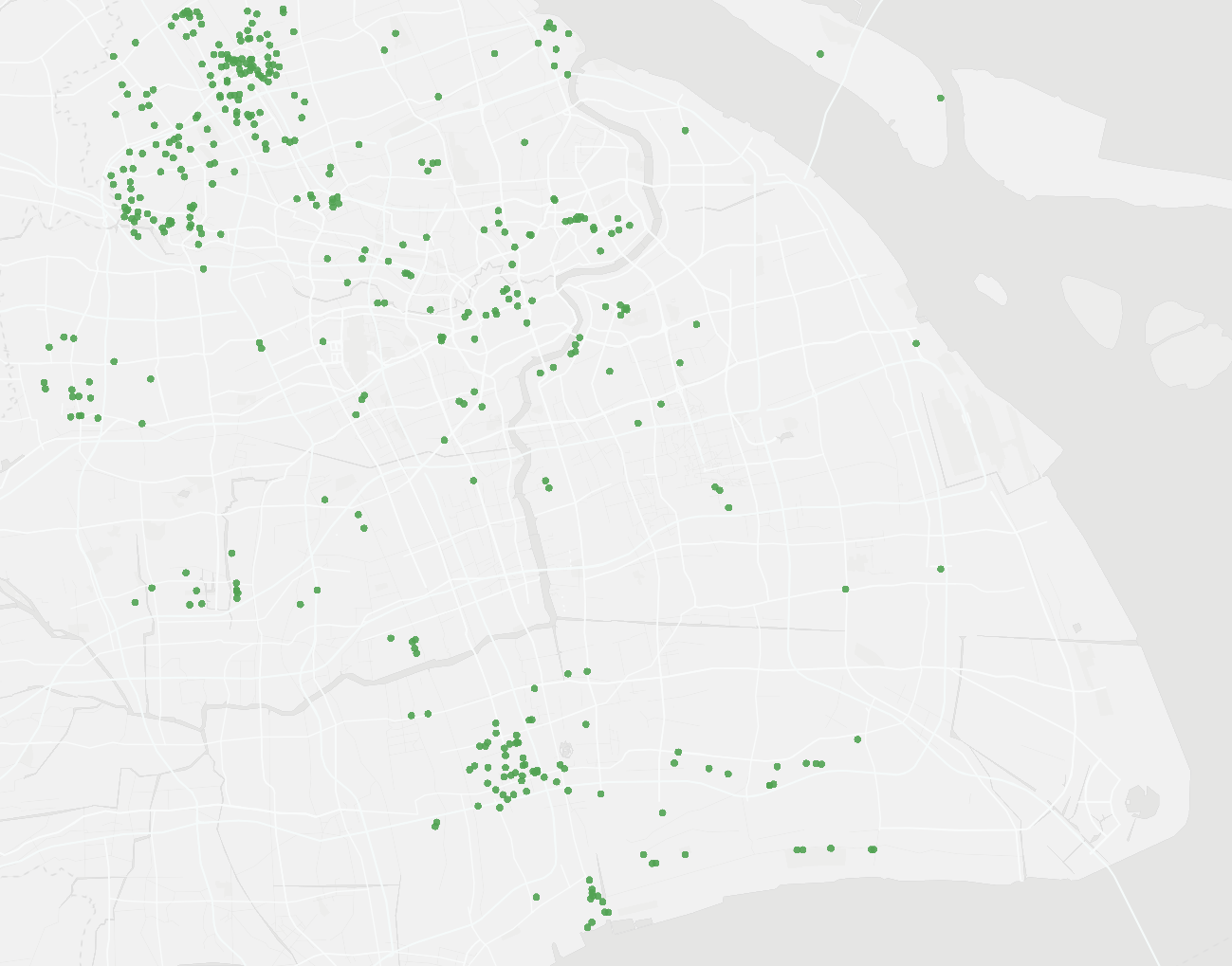}
\caption{Station distribution in Jan.}
\end{subfigure}
\begin{subfigure}{0.5\columnwidth}
\includegraphics[width=\textwidth]{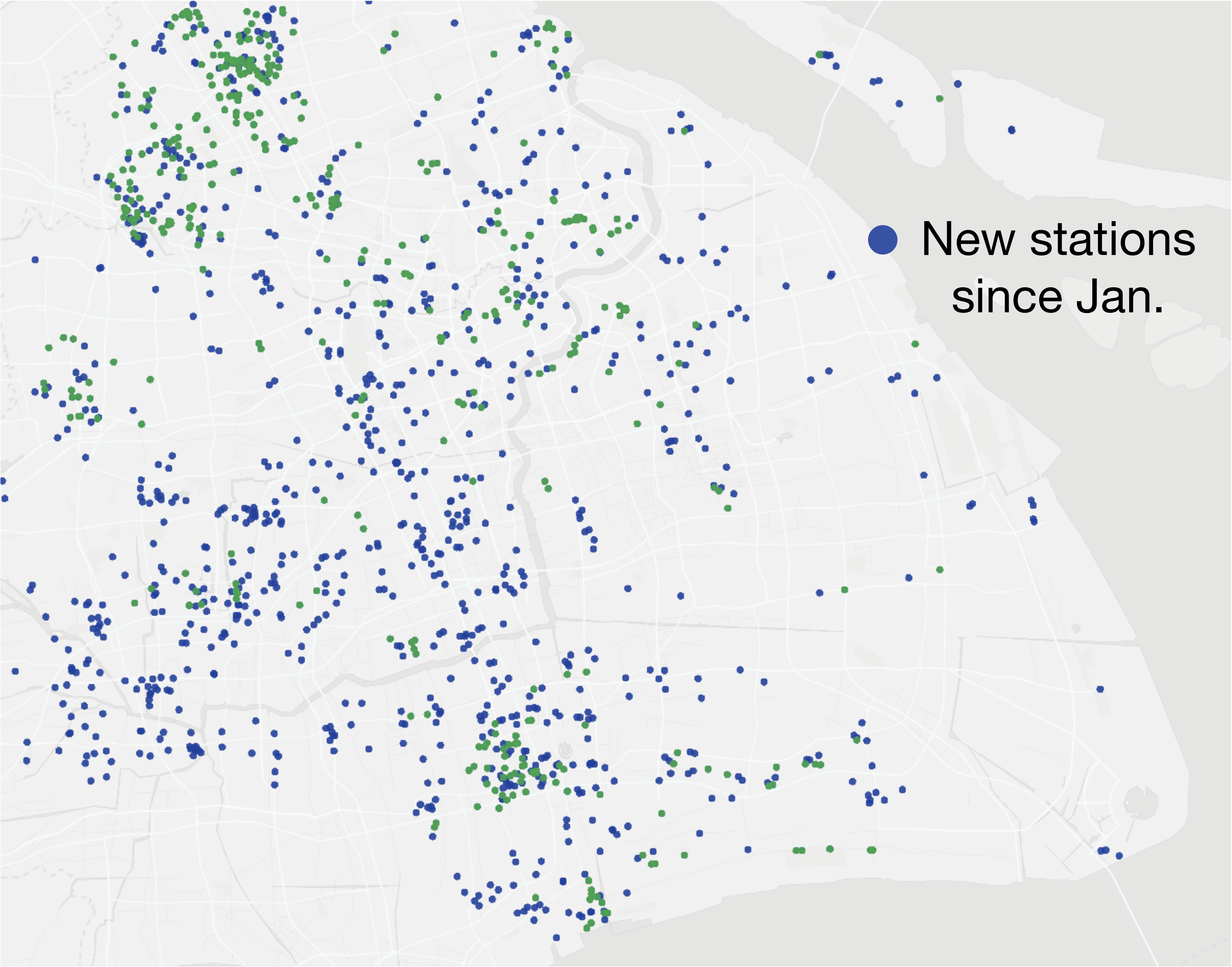}
\caption{Station distribution in Jul.}
\end{subfigure}
\begin{subfigure}{0.5\columnwidth}
\includegraphics[width=\textwidth]{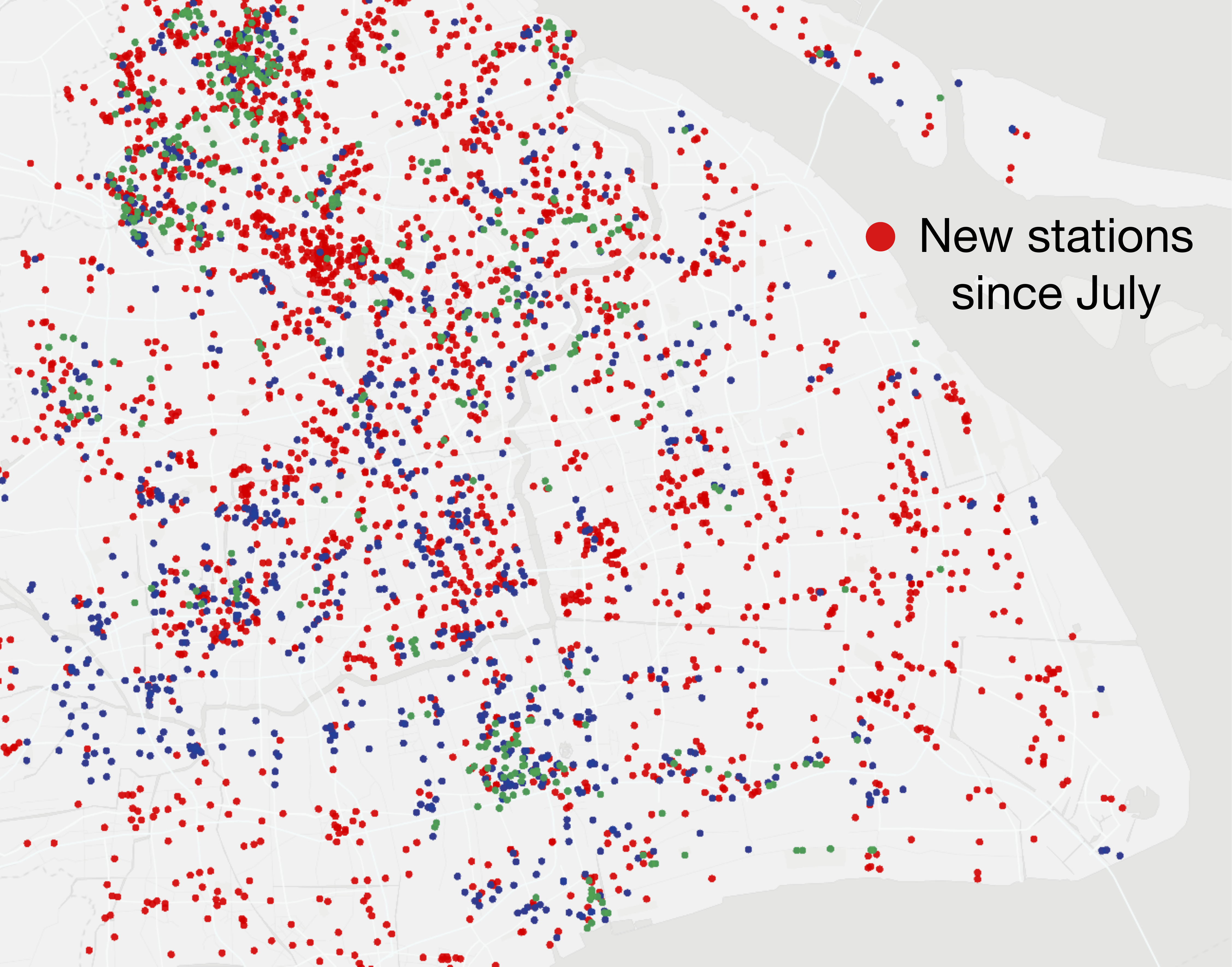}
\caption{Station distribution in Dec.}
\end{subfigure}
\begin{subfigure}{0.5\columnwidth}
\includegraphics[width=\textwidth]{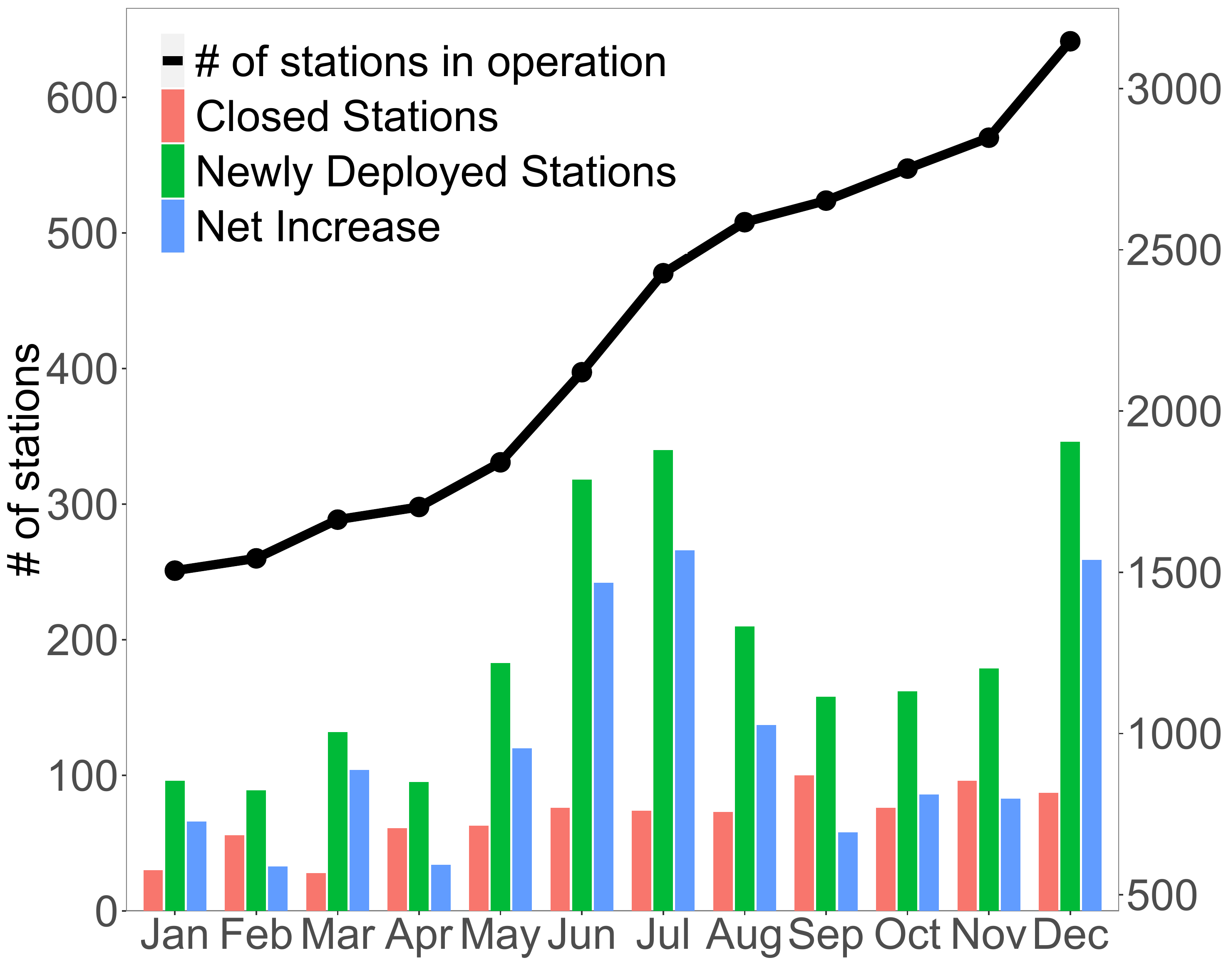}
\caption{Statistics of stations.}
\end{subfigure}
\caption{The expansion process of an EV sharing system in Shanghai during the year 2017. Images better viewed in colour.}
\label{fig:expansion-process}
\end{figure*}

\section{Problem Formulation}
\label{sec:problem}
In this section, we first introduce some key concepts used throughout the paper, then we formulate the problem of dynamic demand prediction and provide an overview of the proposed framework.

\subsection{Preliminaries}
\label{sub:prob_pre}
\noindent \textbf{EV Stations:}
Let $s_i$ be a station in the Electric Vehicle (EV) sharing system. In this paper, we assume $s_i$ can be represented as a tuple $(\boldsymbol{x}_i,m_i)$, where $\boldsymbol{x}_i$ are the coordinates (e.g. latitude and longitude) of $s_i$, and $m_i$ is the number of charging docks within $s_i$. We also assume that for a given $s_i$, we can extract a number of geospatial features based on its location $\boldsymbol{x}_i$, such as nearby Points of Interest (POI) or the distribution of road networks within a certain radius.

% \noindent \textbf{Instant Station Demand:}
% We define the instant demand of station $s_i$ at timestamp $t$ as the rent/return frequency of $s_i$ when it is available, denoted as $d_i(t)$. In this paper the granularity of timestamp $t$ is days, i.e., we focus on daily station demand, but it is straightforward to adopt other time granularity levels in our framework. 

\noindent \textbf{Station Demand:}
For a station $s_i$, the expected demand $\bar{d}_i$ over a period $[t_s,t_e]$ can be defined as the mean $ \bar{d}_i(t_s,t_e) = |t_e - t_s|^{-1}\sum_{t = t_s}^{t_e} d_i(t)$. In practice, we often consider the expected demand from current time $t$ towards the future, and aggregate it according to some index, e.g., days of the week. Without loss of generality, in this paper we denote the future expected demand of station $s_i$ as $\bar{\boldsymbol{d}}_i = [\bar{d}_i^{\text{Mo}},\bar{d}_i^{\text{Tu}}, ..., \bar{d}_i^{\text{Su}}]$ for different days of the week.

\noindent \textbf{Station Network:}
The stations of the EV sharing system can be modelled as a graph $G = 
(S,A)$, where the nodes $s_i\in S$ are stations as defined above. An edge $a_{ij}\in A$ may encode a certain type of correlation between two stations $s_i$ and $s_j$, e.g., the spatial distance between them, or similarity between their POI/road network features. Sec.~\ref{sec:method} will discuss how our approach constructs multiple graphs to capture such inter-station relationships in more details. 

% \noindent \textbf{Station Network Dynamics:}
% Unlike existing work, in this paper we assume the station network is continuously evolving over time. More specifically, let $G_{t-1} = (S_{t-1}, A_{t-1})$ represents the station network at time $t-1$. We assume at time $t-1$, there is an expansion plan to be implemented before time $t$, which shall expand the current station network from $G_{t-1}$ to the \textit{planned network} $G^{\,\text{P}}_{t-1}$. Let's assume during this a set of new stations $S^+$ will be deployed, while existing stations $S^-$ will be removed. If the expansion plan goes through, then at time $t$ the station network $G_{t}$ becomes $G^{\,\text{P}}_{t-1}$, where $G_{t} = (S_{t}, A_{t})$, $S_t= (S_{t-1} -  S^-)  \cup S^+$ and $A_t =(A_{t-1} - \{a_{ij} | s_i\in S^- \text{ or } s_j\in S^- \} )\cup \{a_{ij} | s_i\in S^+ \text{ or } s_j\in S^+ \}$.

\subsection{Demand Prediction}
\label{sub:prob_ddp}
Suppose that at time $t$, we have the previous topology $G_1,...G_{t}$ and demand $D_1,...,D_{t}$ of the station network, where $D_{t} = \{ d_i(t) | s_i\in G_{t} \}$. The demand prediction problem is that given the historical data, for an arbitrary station we aim to estimate both its expected future demand $\hat{\bar{\boldsymbol{d}}}_i$ and the subsequent $k$ instant demand $[\hat{d}_i(t+1), \hat{d}_i(t+2),..., \hat{d}_i(t+k)]$, which minimise the mean square errors with respect to the ground truth $\bar{\boldsymbol{d}}_i$ and $d_i$:
\begin{equation}
    \label{eq:opt_goal}
    \begin{split}
        \delta_{\bar{d}_i} = | \bar{\boldsymbol{d}}_i|^{-1} \|\hat{\bar{\boldsymbol{d}}}_i - \bar{\boldsymbol{d}}_i\|^2 \text{, and }
        \delta_{d_i} & = k^{-1}\sum_{\tau = t+1}^{t+k} \| \hat{d}_i(\tau) - d_i(\tau) \|^2
    \end{split}
\end{equation}
\section{Methodology}
\label{sec:method}

% \begin{figure*}[t]
% \includegraphics[width=\textwidth]{figs/arch.pdf}
% \caption{The workflow of the proposed dynamic demand prediction approach.}
% \label{fig:system-arch}
% \end{figure*}

\subsection{Temporal Encoding}
\label{sub:local_temporal}

Like in many other shared mobility systems, we observe that the demand of stations in the EV sharing platform exhibits strong temporal correlations, as shown later in Fig.~\ref{fig:datasets}(b). For instance, although it fluctuates largely over time, the demand at a station approximates certain periodical patterns at different days across the week. In that sense, exploiting such knowledge can help significantly in estimating the accurate future demand of current stations, which will have a positive knock-on effect when predicting demand for new stations during expansion. However, those demand patterns are typically influenced by multiple complex factors such as weather, air quality and events, and individual stations may react to those factors very differently. Therefore, it is often not optimal to only incorporate the temporal information globally for the station network, but instead in this paper we model such microdynamics at station level.

Concretely, when a station $s_i$ is deployed, we instantiate a local LSTM network which keeps processing its demand records and the additional temporal information available, e.g. weather, days of the week and public holiday/events. In our implementation, we train the LSTMs with shared weights across stations. Then at a later time $t$, the LSTM encodes the station's historical demand $d_{i}(t), d_{i}(t-1),...$ of $s_i$ as well as the auxiliary information into a temporal feature vector $\textbf{f}_{i}(t)$. Moreover, in this paper we also condition $\textbf{f}_{i}(t)$ with a static station feature $\textbf{c}_i$, which describes key attributes of $s_i$ such as its number of available charging docks $m_i$, nearby POIs and environmental characteristics etc. Therefore, $\textbf{f}_{i}(t)$ and $\textbf{c}_i$ carry important local information about individual stations since they started operating, which are then passed on as the input for spatial encoding. 
%Fig.~\ref{fig:system-arch} shows the workflow of the proposed approach, where we see that at each timestamp we maintain a collection of local LSTMs to encode information of individual stations.

\subsection{Spatial Encoding}
\label{sub:dynamic_spatial}

\subsubsection{Constructing Graphs}
\label{ssub:multi_graph}
As discussed in Sec.~\ref{sub:prob_pre}, at a given time $t$ we represent the station network as a graph $G_t = (S_t, A_t)$, where $S_t$ are the set of current stations and $A_t$ is the adjacent matrix describing the pairwise correlations between them. In practice there are often more than one types of correlations, which can't be effectively captured by a single graph. Therefore in this paper we construct multiple graphs to encode the complex inter-station relationships, particularly the \textit{distance graph}, the \textit{functional similarity graph}, and the \textit{road accessibility graph}.

\noindent \textbf{Distance:}
In most cases, we observe that the demand of stations close to each other are highly correlated, e.g. they may be deployed around the same shopping centre, and thus tend to be used interchangeably. We capture such correlations with a distance graph $A^{\text{D}}$, whose elements are the reciprocal of station distance:
\begin{equation}
\label{eq:dist_graph}
  a_{ij}^{\,\text{D}} =  \|\boldsymbol{x}_i - \boldsymbol{x}_j\|_2^{-1} 
\end{equation}
where $\boldsymbol{x}_i$,$\boldsymbol{x}_j$ are the station coordinates, and $\|\cdot\|_2$ is the Euclidean distance. We also set $\diag (A^{\text{D}})$ to 1 to include self loops. 

\noindent \textbf{Functional Similarity:}
Intuitively, stations deployed in areas with similar functionalities should share comparable demand patterns. For instance, stations close to university campuses typically have significantly higher demand during weekends. We characterise the functionalities of stations by considering the distributions of their surrounding POIs. Suppose we have $P$ different categories of POIs in total, and let $\boldsymbol{p}_i$ be the distribution of the $P$ types of POIs within a certain radius of station $s_i$. The functional similarity graph $A^{\text{F}}$ is then defined as:
\begin{equation}
\label{eq:func_graph}
  a_{ij}^{\,\text{F}} =  \simfunc (\boldsymbol{p}_i, \boldsymbol{p}_j) 
\end{equation}
where $\simfunc()\in [0,1]$ is a similarity measure which quantifies the distance between feature vectors. In our experiments, we use the soft cosine function. 

\noindent \textbf{Road Accessibility:}
Another important factor that affects station demand is the accessibility to road networks. Intuitively, stations close to major ring roads, or within areas that have densely connected streets would have higher demand. To model this, we consider the drivable streets in the vicinity of a station $s_i$ as a local road network, containing different types of road segments and their junctions. We exact a feature vector $\boldsymbol{r}_i$ from the local road network, which encodes information such as the road segments density, average junction degree and mean centrality etc. Given those features, the road accessibility graph can be defined with certain similarity function $\simfunc()$: 
\begin{equation}
\label{eq:func_graph}
  a_{ij}^{\,\text{R}} =  \simfunc (\boldsymbol{r}_i, \boldsymbol{r}_j) 
\end{equation}

\subsubsection{ Multi-graph Convolution}
\label{ssub:graph_conv}
At time $t-1$, given the constructed graphs $\boldsymbol{A}_{t-1} = \{A_{t-1}^{\,D},A_{t-1}^{\,F},A_{t-1}^{\,R}\}$ which describe the inter-station relationships, we propose a dynamic multi-graph GCN to fuse such spatial knowledge with local features $\textbf{f}_{i}(t-1)$ and $\textbf{c}_{i}$ computed by the station-level temporal encoding. We perform multi-graph convolution as follows: 
\begin{equation}
\label{eq:graph_conv}
    \boldsymbol{H}_{t-1}^{(l)} = \sigma \bigg( \sum_{A_{t-1} \in \boldsymbol{A}_{t-1}} f(A_{t-1})\boldsymbol{H}_{t-1}^{(l-1)} \boldsymbol{W}_{t-1}^{(l-1)} \bigg)
\end{equation} 
where $\boldsymbol{H}_{t-1}^{l-1}$ and $\boldsymbol{H}_{t-1}^{l}$ are the hidden features in layers $l-1$ and $l$ respectively, while $\boldsymbol{W}_{t-1}^{l-1} \in \mathbb{R}^{U_{l-1} \times U_{l}}$ is the feature transformation matrix learned through end-to-end training. In particular, the input $\boldsymbol{H}_{t-1}^{(0)}$ is the collection of local features computed at individual stations. $f(A_{t-1})$ is a function on graphs $A_{t-1}$, e.g. the symmetric normalized Laplacian~\cite{Kipf:ICLR:2017} or $k$-order polynomial function of Laplacian~\cite{Geng:AAAI:2019}, and $\sigma$ is a non-linear activation function such as ReLU.

\subsection{Demand Prediction}
\label{sub:demand_prediction}

In this paper, we consider the expected demand of station $s_i$ over different days of the week, i.e.  $\bar{\boldsymbol{d}}_i = [\bar{d}_i^{\text{Mo}},\bar{d}_i^{\text{Tu}}, ..., \bar{d}_i^{\text{Su}}]$. To predict $\bar{\boldsymbol{d}}_i$, we plug in a fully connected network to the context vector $\boldsymbol{H}_{t}$, which is trained to output the future expected demand for each station in the network. In Sec.~\ref{sub:eval_results} we will show that in real-world experiments our approach significantly outperforms the existing techniques in prediction accuracy.

\begin{figure*}[ht]
\centering
\begin{subfigure}{0.5\columnwidth}
\includegraphics[width=\textwidth]{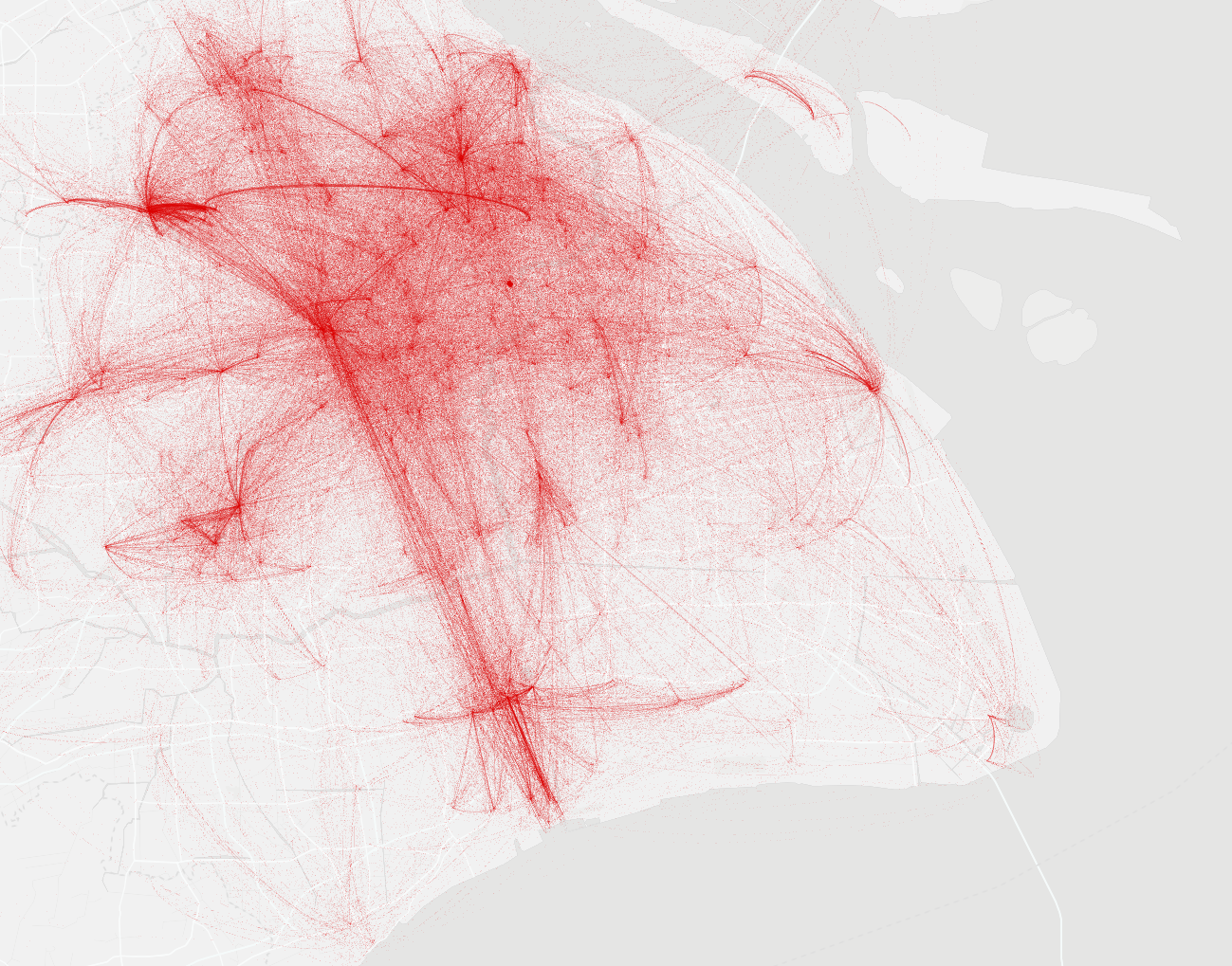}
\caption{}
\label{fig:subim1}
\end{subfigure}
\begin{subfigure}{0.5\columnwidth}
\includegraphics[width=\textwidth]{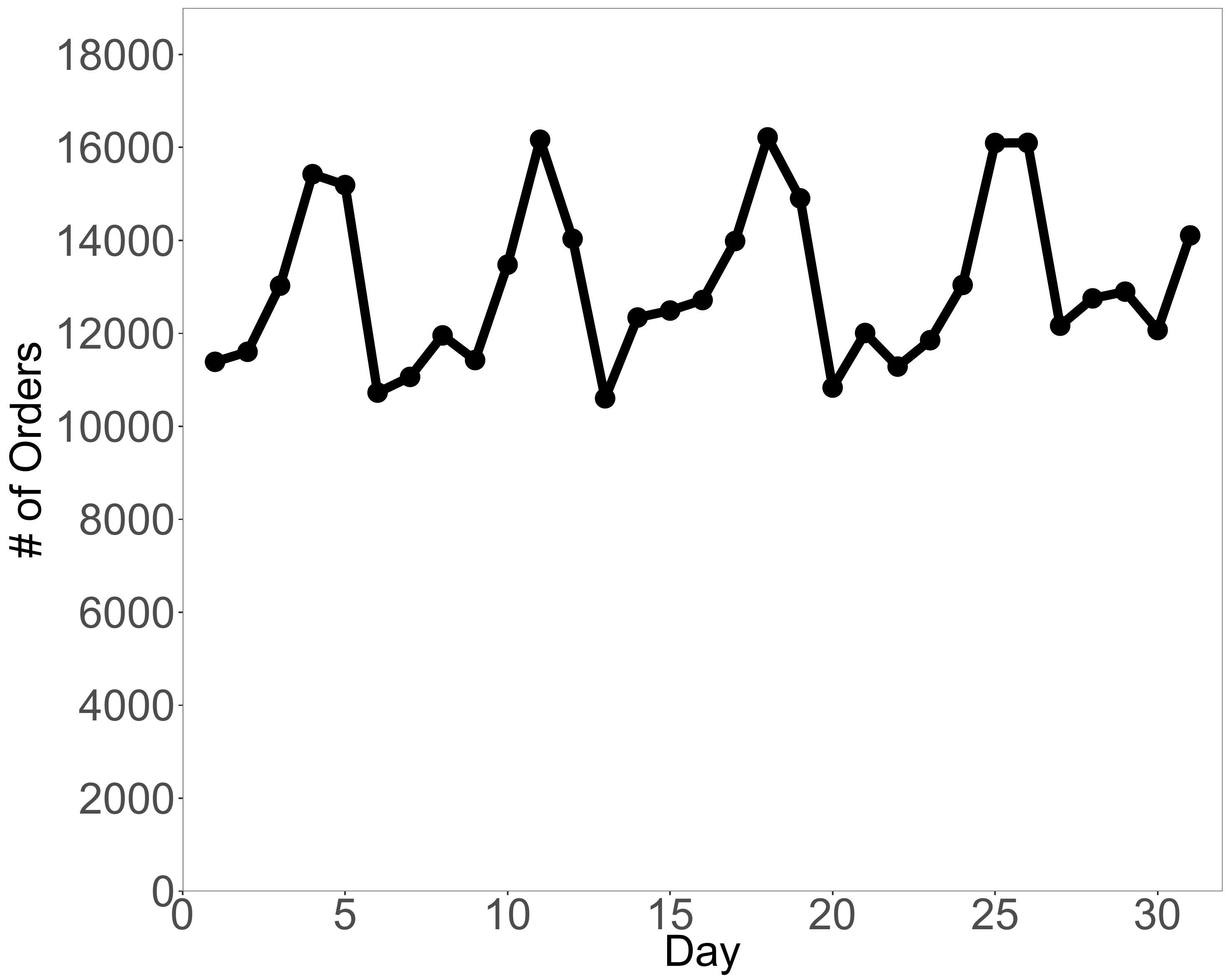}
\caption{}
\label{fig:subim2}
\end{subfigure}
\begin{subfigure}{0.5\columnwidth}
\includegraphics[width=\textwidth]{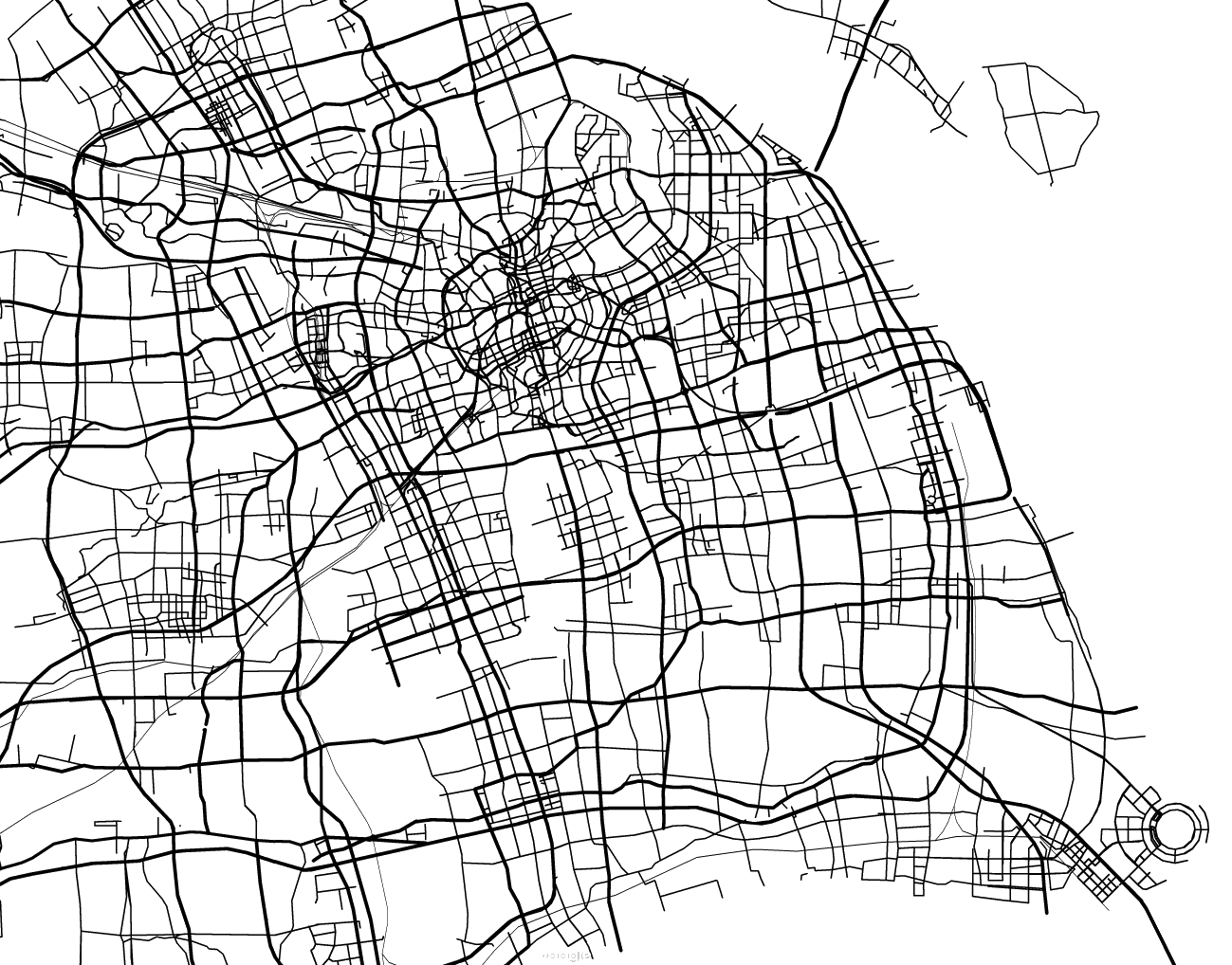}
\caption{}
\label{fig:subim3}
\end{subfigure}
\begin{subfigure}{0.5\columnwidth}
\includegraphics[width=\textwidth]{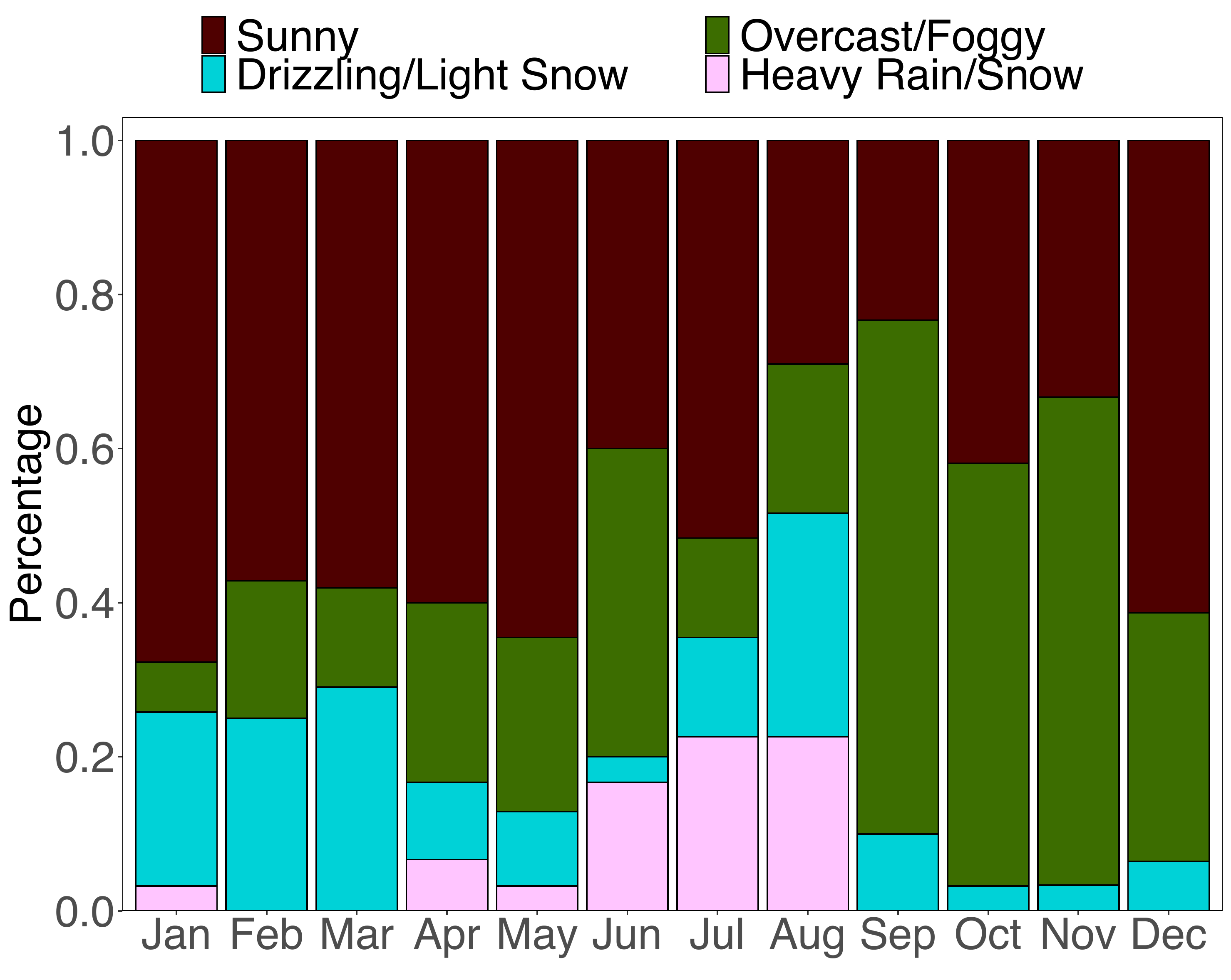}
\caption{}
\label{fig:subim4}
\end{subfigure}
\caption{Visualisation of data used in the experiments. (a) Spatial distribution of orders in one month. (b) Number of orders in one month. (c) Road network in Shanghai. (d) Weather distribution of Shanghai in 2017. }
\label{fig:datasets}
\end{figure*}

\begin{figure*}[t]
\centering
\begin{subfigure}{0.5\columnwidth}
\includegraphics[width=\textwidth]{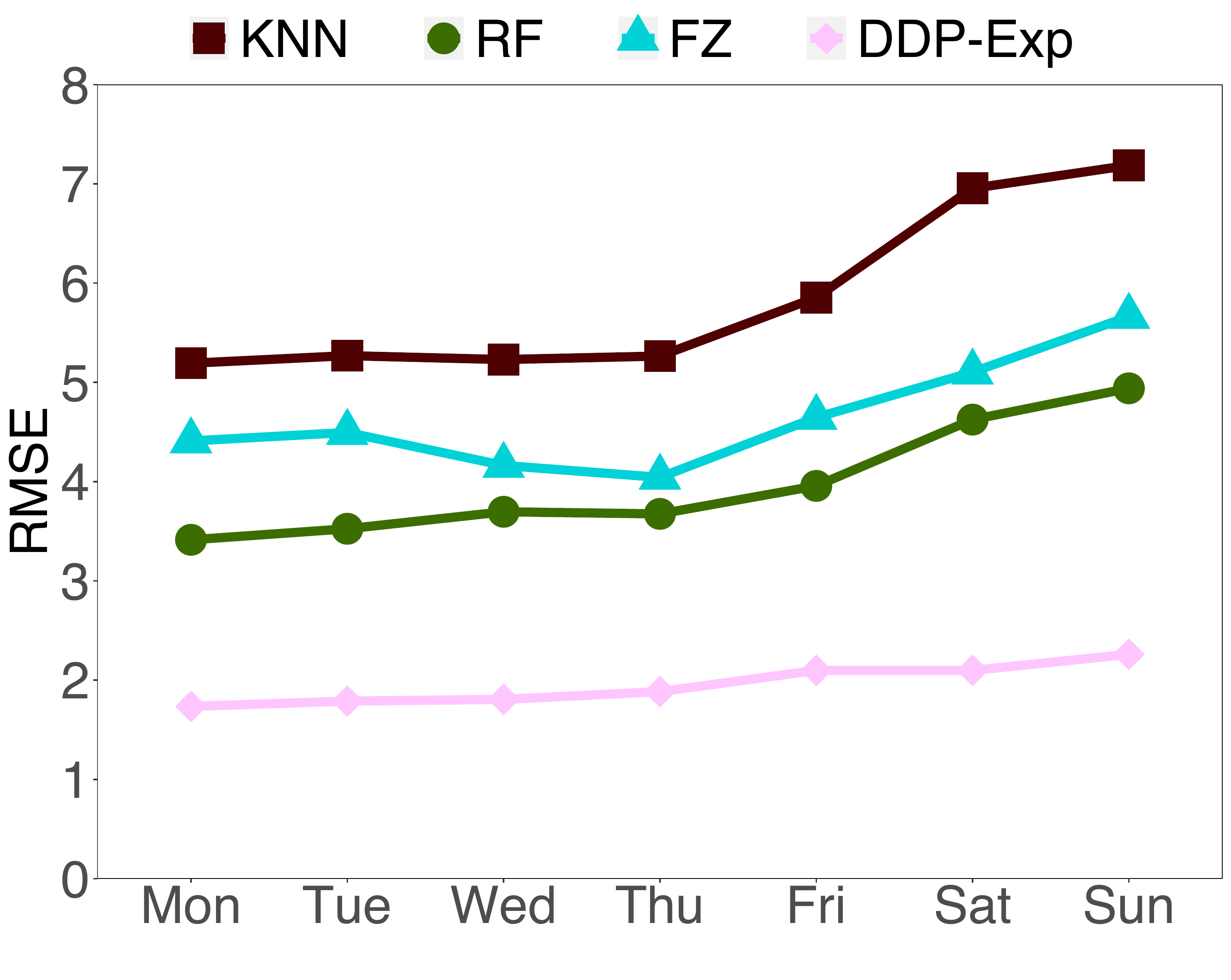}
\caption{}
\end{subfigure}
\begin{subfigure}{0.5\columnwidth}
\includegraphics[width=\textwidth]{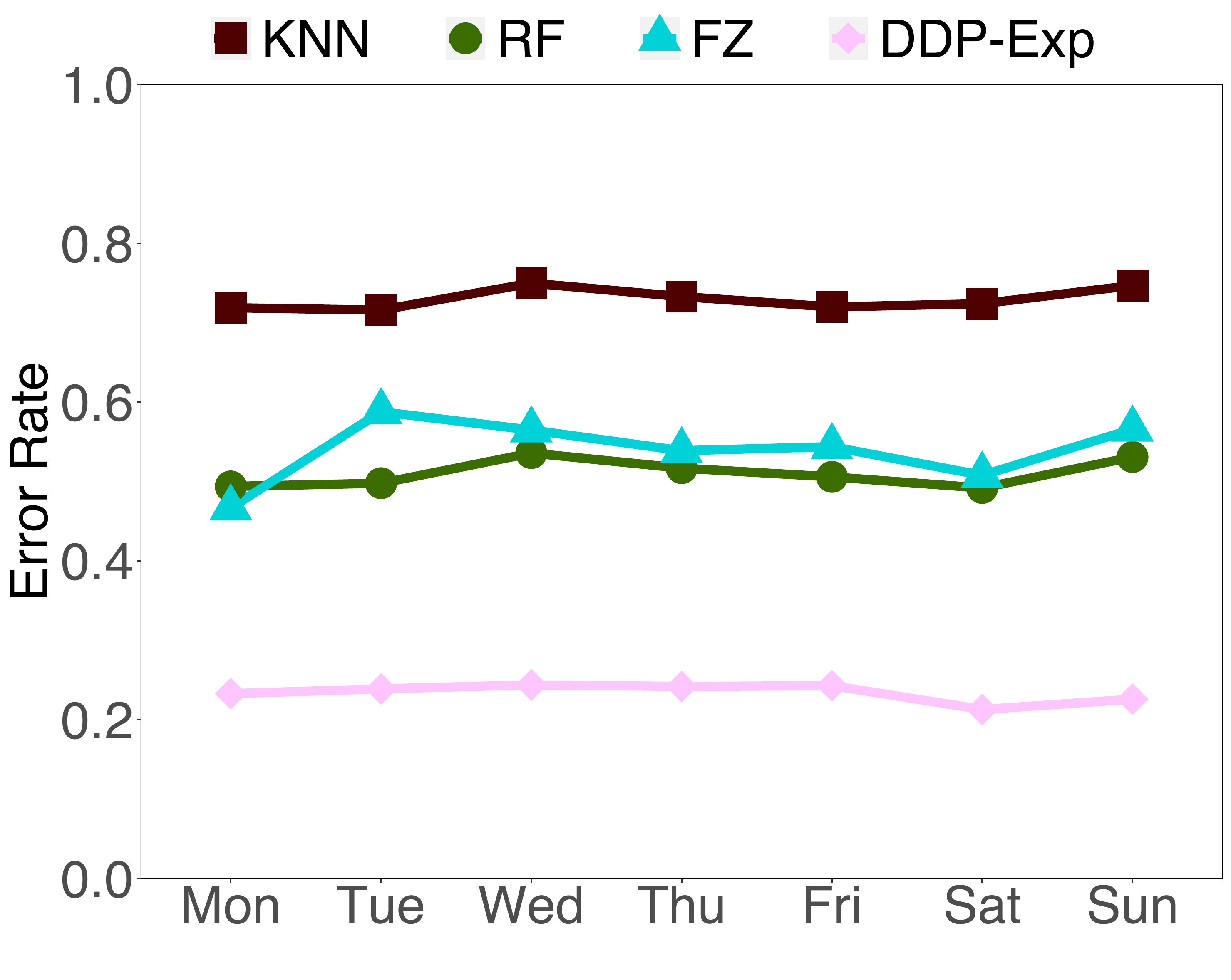}
\caption{}
\end{subfigure}
\begin{subfigure}{0.5\columnwidth}
\includegraphics[width=\textwidth]{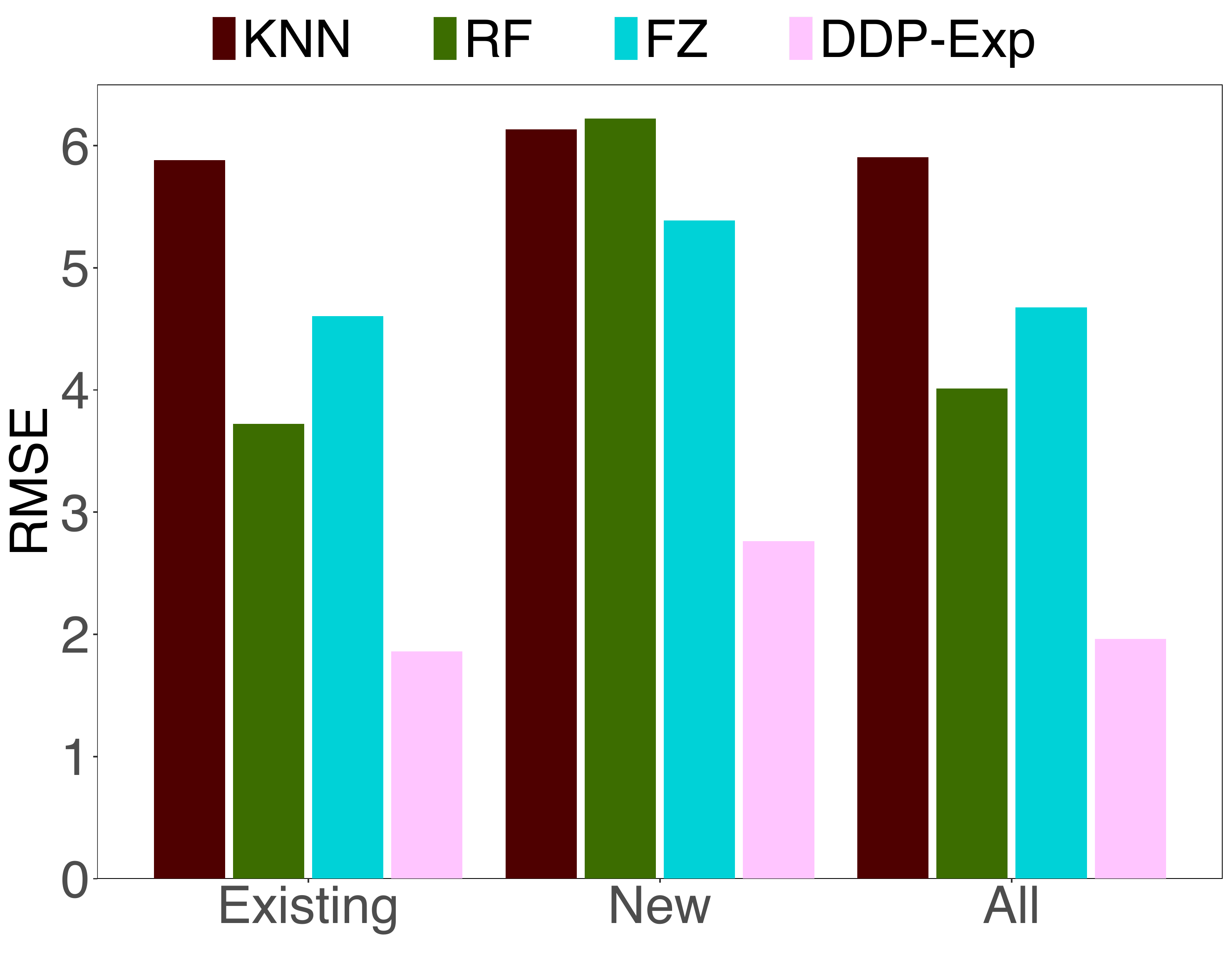}
\caption{}
\end{subfigure}
\begin{subfigure}{0.5\columnwidth}
\includegraphics[width=\textwidth]{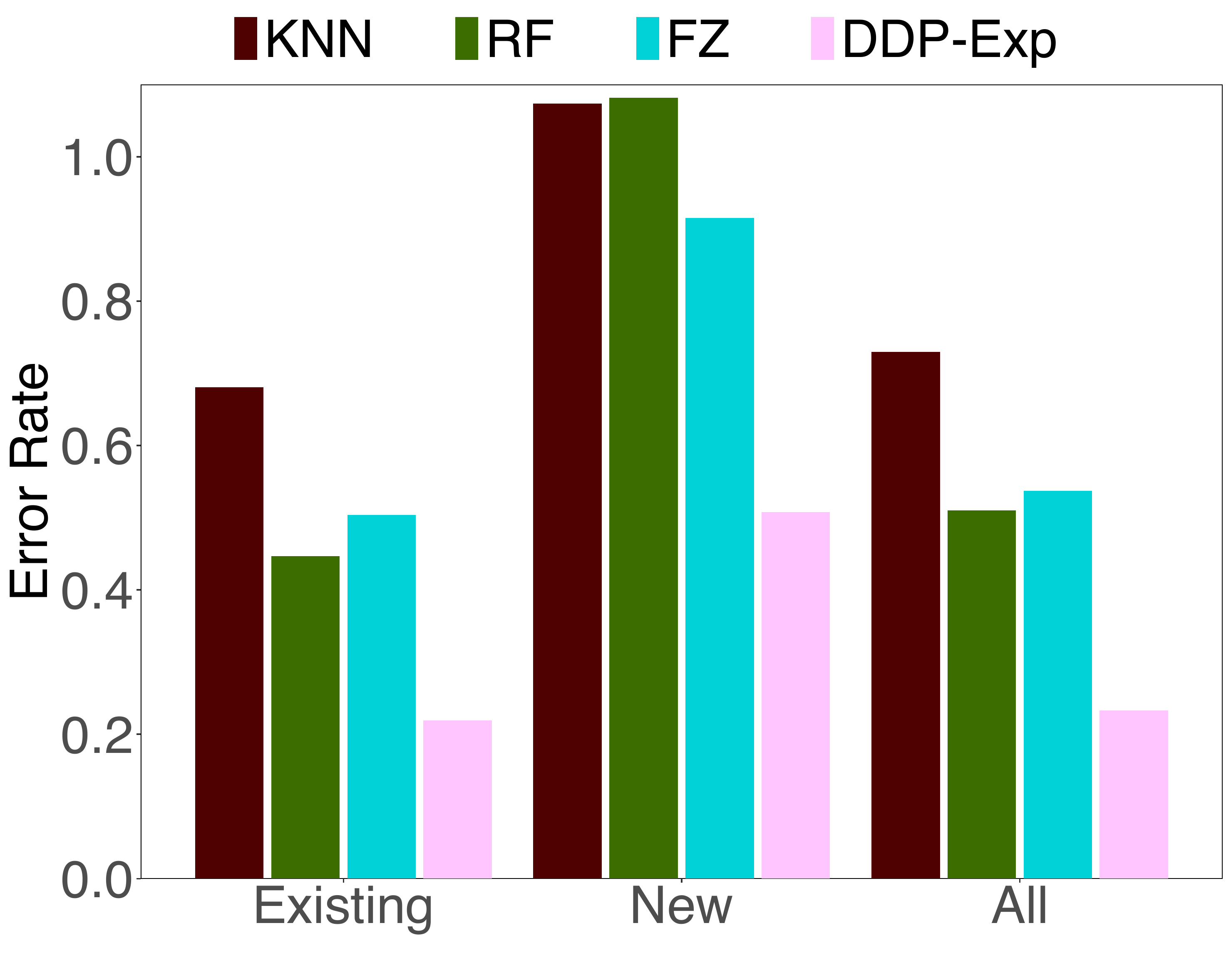}
\caption{}
\end{subfigure}

\caption{Performance on predicting the expected demand. (a) RMSE and (b) ER of all stations across different days in the week. (c) RMSE and (d) ER of existing vs. newly deployed stations vs. all stations averaged over all days of the week.} 
\label{fig:result-exp-demand}
\end{figure*}

\section{Evaluation}
\label{sec:eval}
In this section, we evaluate the performance of the proposed dynamic demand prediction approach on a real electric vehicle sharing platform in Shanghai, China. We describe the datasets and baseline approaches considered in our experiments (Sec.~\ref{sub:eval_data} and~\ref{sub:eval_baselines}), and then discuss the experimental results in Sec.~\ref{sub:eval_results}.  

\subsection{Datasets}
\label{sub:eval_data}

\noindent \textbf{Electric Vehicle (EV) Sharing Data: }
Our EV data is collected from real-world operational records of an EV sharing platform for one year (Jan. to Dec. 2017), containing information on its renting/returning orders, and the detailed expansion process of the station network. In particular, there were 1705 stations and 4725 electric vehicles at the beginning of 2017, while as of Dec 2017 it had 3127 stations with a fleet of 16148 vehicles in operation. In total, the raw data contains 6,843,737 records, which were generated by approximately 0.36 million users. Fig.~\ref{fig:datasets}(a) visualises the spatial distribution of the orders (represented as lines between pick up and return stations) in a month. Fig.~\ref{fig:datasets}(b) shows the number of orders in different days over a month, which exhibits clear periodic patterns with peaks in weekends.  

\noindent \textbf{POI Data: }
We also collect the Point Of Interest (POI) date from an online map service provider in China. In total we have extracted 4,126,844 POI entries in Shanghai, each of which consists of a GPS coordinate and a category label. In our experiments, for each station we consider the POIs within 1km radius. 

\noindent \textbf{Road Network Data: }
We extract road network data in Shanghai using OSMnx~\cite{Boeing:CEUS:2017} from OpenStreetMap, which is formatted as a graph (visualised in Fig.~\ref{fig:datasets}(c)). Similar with the POIs, we consider the subgraphs within 1km radius of the stations. In our data, on average a subgraph contains road segments of length 13.85km and approximately 39 junctions, with a mean degree of 4.28.

\noindent \textbf{Meteorology Data: }
Finally, we collect the daily weather data in Shanghai for 2017 from the publicly available sources. Each record describes weather conditions of the day, which falls into four different categories: \textit{sunny}, \textit{overcast/foggy}, \textit{drizzling/light snow} and \textit{heavy rain/snow}. Fig.~\ref{fig:datasets}(d) shows the distribution of weather conditions in Shanghai over the 12 months.

% \begin{table}[t]
% \begin{tabular}{|l|r|l|r|}
% \hline
% \multicolumn{1}{|c|}{POI Type} & \multicolumn{1}{c|}{Number} & \multicolumn{1}{c|}{POI Type} & \multicolumn{1}{c|}{Number} \\ \hline
% Hospitals                      & 4745                        & Banks                         & 2988                        \\ \hline
% Tourist attractions            & 2696                       & Companies                     & 89,747                      \\ \hline
% Gov. organizations              & 16,425                       & Higher education              & 6922                        \\ \hline
% Airport services               & 126                         & Residences                    & 51,089                      \\ \hline
% Subway stations                & 1,729                        & Hotels                        & 18,234                       \\ \hline
% Bus stations                   & 41,475                       & \multicolumn{1}{c|}{...}        & \multicolumn{1}{c|}{...}      \\ \hline
% \end{tabular}
% \caption{Statistics of some POI categories in our data.}
% \label{tbl:POI}
% \end{table}

\subsection{Baselines and Metric}
\label{sub:eval_baselines}
% We evaluate two variants of the proposed dynamic demand prediction approach respectively: 1) \textbf{DDP-Exp}, which predicts the future \textit{expected demand} of stations; and 2) \textbf{DDP-Seq}, which forecasts the \textit{instant demand} of stations in a subsequent time window. Both of the two variants share the same local temporal and dynamic spatial encoding processes, but they implement the two different branches in demand prediction (as discussed in Sec.~\ref{sub:demand_prediction}).

In particular, we compare our approach DDP-Exp with the following baselines:

% TODO: How to predict existing, new?
\noindent \textbf{KNN}, which uses a linear regressor to predict the expected demand of existing stations. For the planned stations, it estimates their demand with standard KNN, based on the similarity of features (e.g. POIs) between them and the existing stations.  

\noindent \textbf{Random Forest (RF)}, which shares the similar idea as KNN, but trains a random forest as the predictor.

\noindent \textbf{Functional Zone (FZ)}, which implements the state of the art demand prediction approach for system expansion in~\cite{Liu:KDD:2017}. Note that we don't have taxi records in our data, but instead we directly feed the ground truth check-in/out to favour this approach.

% For DDP-Seq which computes the instant demand, we consider three competing algorithms:  

% \noindent \textbf{ARIMA + KNN}, which uses Auto-Regressive Integrated Moving Average (ARIMA)~\cite{Williams:JTE:2003} to forecast multi-step demand at existing stations, and then uses KNN to estimate demand at new station based on station features such as POIs. 

% \noindent \textbf{LSTM + KNN}, which is similar with A-KNN, but trains LSTM networks for temporal modelling.

% \noindent \textbf{Multi-graph GCN (MGCN)}, which implements a similar framework as the state of the art in~\cite{Chai:SIGSPATIAL:2018}. To perform fair comparison, here we use our dynamic multi-graph GCN implementations that can handle new/closed stations, and consider the same data sources as in our approach.

For all approaches, we adopt the Root Mean Squared Error (RMSE) and the Error Rate (ER) as the performance metric:
\begin{equation}
    \label{eq:metric}
    \begin{split}
    RMSE = \sqrt{ \frac{1}{N}\sum_{i=1}^N (\hat{z}_i - z_i)^2} \text{, and }
    ER   = \frac{\sum_{i=1}^N |\hat{z}_i - z_i|}{\sum_{i=1}^N z_i}
    \end{split}
\end{equation}
where $\hat{z}_i$ and $z_i$ are predicted and ground truth values respectively.

%% How to actually predict after training? Need to eat some data before right? Exp and Instant?
% We implement the deep neural networks in the proposed approach with TensorFlow 1.10.0, and use the Adam optimiser with learning rate of 0.001. The networks are trained on a single Titan X GPU from scratch. For all approaches, we randomly select two months of data for training while the subsequent month for testing, and report the average performance.

\subsection{Evaluation Results}
\label{sub:eval_results}

% \begin{figure}[t]
% \centering
% \begin{subfigure}{0.5\columnwidth}
% \includegraphics[width=\textwidth]{figs/instant-rmse.pdf}
% \caption{}
% \end{subfigure}~
% \begin{subfigure}{0.5\columnwidth}
% \includegraphics[width=\textwidth]{figs/instant-er.pdf}
% \caption{}
% \end{subfigure}
% \caption{Performance on predicting the instant demand. (a) RMSE and (b) ER of the competing approaches.} 
% \label{fig:result-inst-demand}
% \end{figure}

% \begin{figure}[t]
% \centering
% \begin{subfigure}{0.5\columnwidth}
% \includegraphics[width=\textwidth]{figs/days_algorithms_rmse.pdf}
% \caption{}
% \end{subfigure}~
% \begin{subfigure}{0.5\columnwidth}
% \includegraphics[width=\textwidth]{figs/days_algorithms_er.pdf}
% \caption{}
% \end{subfigure}
% \caption{(a) RMSE and (b) ER of the predicted instant demand for different prediction lengths.} 
% \label{fig:result-acc-len}
% \end{figure}

% \begin{figure}[t]
% \centering
% \begin{subfigure}{0.5\columnwidth}
% \includegraphics[width=\textwidth]{figs/dropout_rmse.pdf}
% \caption{}
% \end{subfigure}~
% \begin{subfigure}{0.5\columnwidth}
% \includegraphics[width=\textwidth]{figs/dropout_er.pdf}
% \caption{}
% \end{subfigure}
% \caption{Sensitivity of our approach with different levels of augmented station network dynamics.} 
% \label{fig:result-p}
% \end{figure}

\noindent \textbf{Accuracy of Predicting Expected Demand: }
The first set of experiments evaluate the overall accuracy when predicting the expected demand of stations. Fig.~\ref{fig:result-exp-demand}(a) and (b) show the RMSE and ER of the proposed approach (DDP-Exp) and competing algorithms over different days of the week. We see that comparing to naive KNN, the random forest based approach (RF) can reduce the RMSE by about 30\% while ER by 20\%. However, our approach (DDP-Exp) performs significantly better, and can achieve up to three times improvement in both RMSE and ER. In particular, on average the RMSE of DDP-Exp is approximate 1.961, which means when predicting the station's expected demand, the value estimated by our approach is only about $\pm$2 with respect to the ground truth. This confirms that the proposed approach can effectively model the complex temporal and spatial dependencies within the evolving station network, and exploits that to make more accurate predictions. In addition, we observe that the RMSE tends to increase on weekends compared to weekdays for all algorithms. This is because in practice the absolute demand on weekends is larger, which often leads to bigger RMSE. Note that the ER remains relatively consistent across different days. 

\noindent \textbf{Planned vs. Existing Stations: }
This experiment investigates the prediction performance of different approaches on the planned stations which haven't been deployed yet, and existing stations which have already been in operation. Fig.~\ref{fig:result-exp-demand}(c) and (d) show the average RMSE and ER of the proposed approach (DDP-Exp) and the competing algorithms on the planned, existing, and all stations respectively. We see that all of approaches perform better on the existing stations than the planned. This is expected because for existing stations we have access to their historical demand data, which is not available for planned stations. We also observe that although the functional zone based approach (FZ) performs better than the baselines for the planned stations, it fails on the existing stations (performs worse than RF). This is because by design FZ is tuned to predict demand of new stations in the context of system expansion, but not for existing ones. Finally, we see that for both planned and existing stations our approach (DDP-Exp) performs consistently the best. For the planned stations, it halves the errors comparing to the state of the art approach FZ, while for the existing stations, it offers about three-fold improvement over the baselines.

\section{Related Work}
\label{sec:related}
\noindent \textbf{Demand Prediction for Shared Mobility:}
Predicting user demand in shared mobility services (e.g. taxi and bike- or vehicle-sharing systems) has received considerable interest in various research communities. Most of the existing work takes the historical usage (e.g. picking-up and returning records), geospatial data such as POIs, and other auxiliary information (e.g. weather) into account, and builds prediction models that can forecast demand over certain periods or aggregated time slots. They also predict the demand at different spatial granularity, e.g. over the entire systems~\cite{Yin:Stanford:2014, Wang:Thesis:2016}, grids/regions~\cite{Geng:AAAI:2019}, station clusters~\cite{Froehlich:IJCAI:2009,Li:SIGSPATIAL:2015,Mahony:AAAI:2015}, or individual stations~\cite{Hulot:KDD:2018,Chai:SIGSPATIAL:2018,Liu:ICDM:2015,Yang:MOBISYS:2016,Zeng:LSTS:2016}. This paper falls into the last category since we aim to predict station-level demand of EV sharing platforms. However, our work is fundamentally different in that we assume the station network is not static, but dynamically evolving, i.e. stations can be deployed or closed at arbitrary times. In this case, state of the art station-level demand predictors (e.g.~\cite{Hulot:KDD:2018}) will fail because they rely heavily on station historical data to make predictions, which are not available for newly deployed stations.  

\noindent \textbf{Shared Mobility Expansion:}
There is also a solid body of work focusing on modeling the expansion process of shared mobility systems, e.g. planning for optimal new stations~\cite{Xiong:IJCAI:2015,Liu:ICDM:2015}, or increasing the capacity of existing stations~\cite{Du:KDD:2018}. However, all of them assume that demand of the stations (renting and returning) are known, or can be estimated from other data sources such as taxi records, which is different from our work. On the other hand, the work in ~\cite{Liu:KDD:2017} proposes a functional zone based hierarchical demand predictor for shared bike systems, which can estimate the average demand at newly deployed stations across different expansion stages. Our work shares similar assumptions with~\cite{Liu:KDD:2017}, yet differs substantially: 1) instead of fixed stages, we can predict demand while the entire station network is dynamically expanding; 2) we are able to estimate both the instant and expected demand of new or existing stations, while~\cite{Liu:KDD:2017} can only predict aggregated demand patterns; and finally 3) we don't require historical mobility data in the newly expanded areas, like the taxi trip records used in~\cite{Liu:KDD:2017}.

\noindent \textbf{Graph-based Deep Learning: }
Due to their non-Euclidean nature, many real-world problems such as demand/traffic/air quality forecasting that require spatio-temporal analysis have been tackled with the emerging graph-based deep learning techniques~\cite{Li:ICLR:2018, Yao:AAAI:2018,Geng:AAAI:2019,Chai:SIGSPATIAL:2018}. In particular, existing work often employs the graph convolutional neural network~\cite{Bruna:arXiv:2013} to capture the spatial correlations, where temporal dependencies are typically modelled with recurrent neural networks. For instance,~\cite{Li:ICLR:2018} models the traffic flow as a diffusion process on directed graphs for traffic forecasting, while~\cite{Yao:AAAI:2018} and~\cite{Geng:AAAI:2019} propose frameworks that use multi-graph convolutional neural networks (CNNs) to predict demand for taxi and ride-hailing services. Another work in~\cite{Chai:SIGSPATIAL:2018} uses an encoder-decoder structure on top of multi-graph CNNs to estimate flow between stations in bike sharing systems, which bears a close resemblance to this paper. However, unlike~\cite{Chai:SIGSPATIAL:2018} who only output demand at the immediate next timestamp, our work considers a sequence to sequence model with attention mechanism to perform multi-step forecasting towards future demand. In addition, none of the above approaches can work on new stations where historical data is not available.

\section{Conclusion}
\label{sec:Conclusion}
In this paper, we propose a novel demand prediction approach for electric vehicle (EV) sharing systems, which learns the complex system dynamics, and is able to robustly predict demand for stations. Specifically, we first encode the local temporal information at individual station level, and then fuse the extracted features with graph convolutional neural networks (GCN) to account for the spatial dependencies between stations. The demand of stations is estimated by a prediction network, which forecasts the long-term expected demand of the system. We evaluate our approach on data collected from a real-world EV sharing platform for a year. Extensive experiments have shown that our approach consistently outperforms the state of the art in predicting demand of the EV sharing system.

\bibliographystyle{ACM-Reference-Format}
\bibliography{main}

\end{document}